\documentclass[letterpaper]{article} 
\usepackage[submission]{aaai25}  
\usepackage{times}  
\usepackage{helvet}  
\usepackage{courier}  
\usepackage{multirow}
\usepackage{booktabs}
\usepackage[hyphens]{url}  
\usepackage{graphicx} 
\usepackage{natbib}  
\usepackage{caption} 
\usepackage{placeins}
\usepackage{algorithm}
\usepackage{algorithmic}
\usepackage{amssymb}
\usepackage{amsfonts}
\usepackage{amsmath}
\usepackage{newfloat}
\usepackage{listings}
\DeclareCaptionStyle{ruled}{labelfont=normalfont,labelsep=colon,strut=off} 
\floatstyle{ruled}
\newfloat{listing}{tb}{lst}{}
\floatname{listing}{Listing}
\title{AutoMixQ: Self-Adjusting Quantization for High Performance Memory-Efficient Fine-Tuning}

\usepackage{bibentry}

\begin{document}

\maketitle

\begin{abstract}
Fine-tuning large language models (LLMs) under resource constraints is a significant challenge in deep learning. Low-Rank Adaptation (LoRA), pruning, and quantization are all effective methods for improving resource efficiency. However, combining them directly often results in suboptimal performance, especially with uniform quantization across all model layers. This is due to the complex, uneven interlayer relationships introduced by pruning, necessitating more refined quantization strategies. To address this, we propose AutoMixQ, an end-to-end optimization framework that selects optimal quantization configurations for each LLM layer. AutoMixQ leverages lightweight performance models to guide the selection process, significantly reducing time and computational resources compared to exhaustive search methods. By incorporating Pareto optimality, AutoMixQ balances memory usage and performance, approaching the upper bounds of model capability under strict resource constraints. Our experiments on widely used benchmarks show that AutoMixQ reduces memory consumption while achieving superior performance. For example, at a 30\% pruning rate in LLaMA-7B, AutoMixQ achieved 66.21\% on BoolQ compared to 62.45\% for LoRA and 58.96\% for LoftQ, while reducing memory consumption by 35.5\% compared to LoRA and 27.5\% compared to LoftQ.

\end{abstract}

\section{Introduction}
The advent of large language models (LLMs) has revolutionized various natural language processing (NLP) tasks, such as machine translation \cite{zhang2023prompting, sato2020vocabulary}, sentiment analysis \cite{zhang2023enhancing, deng2023llms}, and speech recognition \cite{min2023exploring}. Despite their impressive capabilities, the resource consumption required to obtain a fine-tuned model suitable for specific tasks remains substantial due to the large number of parameters and high computational demands of LLMs \cite{frantar2023sparsegpt}. To address these issues, various compression techniques, including pruning \cite{ma2023llm, xia2023sheared}, quantization \cite{shao2023omniquant, lee2023enhancing}, and distillation \cite{gu2023minillm, tan2023gkd}, have been proposed. These methods have shown that compressing large models can lead to better performance at a lower cost compared to direct training.

Compared to methods that directly compress model parameters, another memory-efficient fine-tuning method is Low-Rank Adaptation (LoRA) \cite{hu2021lora}, which can efficiently add additional knowledge while retaining the original model's capabilities. Previous works \cite{ma2023llm, xia2023sheared} have logically combined these two approaches. They first use structured pruning to remove less important parameters from the model and then apply LoRA fine-tuning to quickly recover performance. The resulting compressed model closely approximates the performance of the pre-pruned model. Recent studies, including QLoRA \cite{dettmers2024qlora} and LoftQ \cite{li2023loftq}, have adopted quantization techniques during LoRA fine-tuning to further reduce memory consumption, but they have not combined these techniques with pruning. To achieve further memory savings by combining pruning and quantization in LoRA fine-tuning, one must address the suboptimality of using fixed configurations across all layers, as existing research \cite{zhang2023adalora} shows that layer importance varies. Moreover, pruning disrupts the originally uniform computational complexity of each layer, making the interlayer relationships in the model more complex.

In quantization\citep{peri2020deploying, liu2023llm, frantar2022gptq, xiao2023smoothquant}, higher quantization precision corresponds to lower errors. Utilizing this property, we can effectively allocate computational resources to different layers through mixed-precision quantization without requiring additional memory. In other words, layer-wise mixed-precision quantization adapts to the complex interlayer relationships mentioned earlier. However, if the quantization configuration for each layer varies across different models and tasks, it results in countless combinations and a vast search space. Additionally, testing the performance of each quantization configuration in practice is highly impractical due to the time-consuming nature of the process. To address the challenge of quickly determining an appropriate quantization configuration and achieving an optimal balance between performance and memory consumption, we propose an innovative end-to-end automatic optimization process, AutoMixQ. This learning-based approach leverages a lightweight performance model and Pareto optimality \cite{censor1977pareto} to efficiently explore the search space and determine the best quantization precision allocation to balance performance and memory usage. Our goal is to approximate the upper bounds of model performance under resource constraints. 

The main contributions are summarized as follows:

\begin{itemize}
\item We combine pruning, quantization, and LoRA in a complementary way, exploring the upper bounds of fine-tuned model performance under resource constraints, and formalizing it as an optimization problem.
\item We propose an end-to-end automatic optimization process that uses lightweight performance model and the Pareto optimality to self-adjust quantization configurations during fine-tuning, bringing the model closer to the performance upper bounds.
\item Extensive experiments on widely-used benchmarks demonstrate that AutoMixQ consistently outperforms both LoRA and LoftQ in terms of memory usage while achieving comparable or better task performance. For instance, at a 50\% pruning rate in LLaMA-7B, AutoMixQ attains 53.82\% on BoolQ compared to 51.80\% for LoftQ and 43.76\% for LoRA, while saving up to 7.68 GB of memory compared to LoRA. Similarly, at a 30\% pruning rate, AutoMixQ achieves 66.21\% on BoolQ versus 62.45\% for LoRA and 58.96\% for LoftQ, saving up to 8.13 GB of memory compared to LoRA.
\end{itemize}

\section{Background and Motivation}

\subsection{Low-Rank Adaptation (LoRA)}\label{sec-bkg-lora}

For a LLM consisting of $n$ layers, the weight matrices at each layer, denoted as $W$, undergo updates through an update matrix $\Delta W$. This update matrix is decomposed into the product of two low-rank matrices $A$ and $B$, where $A \in \mathbb{R}^{d \times r}$ and $B \in \mathbb{R}^{r \times d}$, with $r$ being the rank, a hyperparameter fixed across all layers. In this approach, the original weight matrix $W$ remains frozen, while only $\Delta W$, represented by the product $AB$, is updated.
The forward computation can be expressed as:
\begin{equation}
\small
f(x) = (W + \Delta W)X + b = (WX + b) + (AB)X.
\end{equation}
Given the rank $r$ is typically much smaller than the dimension $d$, the number of parameters is significantly reduced from $d^2$ to $2dr$. This optimization can reduce the trainable parameters during the learning.

\subsection{Quantization}
\textbf{Quantization.} Quantization is an essential technique used to reduce the computational and memory overhead of large-scale models by converting high-precision numerical values, such as a 32-bit floating-point number $X^{\text{HP}} \in \mathbb{R}$, into a lower-bit integer representation $X^{\text{INT}} \in \{0, 1, \dots, 2^{N}-1\}$. This process is mathematically expressed as:
\begin{align}
\label{eq:quant}
    X^{\text{INT}} = \text{round}\left((2^{N} - 1) F\left( X^{\text{HP}} \right) \right),
\end{align}
where $F(\cdot) \colon \mathbb{R} \rightarrow [0, 1]$ is a normalization function. A typical method is uniform quantization, where $F(X)$ is defined as $F(X) = \frac{X - X_{\min}}{X_{\max} - X_{\min}}$. An alternative approach introduced by QLoRA \citet{dettmers2024qlora} is 4-bit NormalFloat Quantization (NF4), which assumes that the data follows a normal distribution $X \sim \mathcal{N}(0, \sigma^2)$ and applies $F(X) = \Phi(X/\sigma)$, with $\Phi(\cdot)$ representing the cumulative distribution function of a standard normal distribution.

\textbf{Dequantization.} To recover the high-precision values from their quantized forms, a lookup table $\mathcal{T}$ is used, which is defined as:
\begin{align}
\label{eq:lookup_table}
    \mathcal{T}[i] = F^{-1}\left( \frac{i}{2^N - 1} \right), \quad i = 0, 1, \dots, 2^{N}-1,
\end{align}
allowing the integer $X^{\text{INT}}$ to be mapped back to its simulated high-precision counterpart $X^{\text{D}} \in \mathbb{R}$. The dequantization process can be represented as:
\begin{align}
\label{eq:dequant}
    X^{\text{D}} = \mathcal{T}[X^{\text{INT}}].
\end{align}

\textbf{Simulated Quantization for Matrices.} In practice, it is often more efficient to use simulated quantization for matrices rather than directly operating on quantized values \citep{bai2020binarybert, shen2020q}. In this method, quantized weight matrices are stored as encoded integers and are temporarily dequantized into simulated high-precision matrices during multiplication operations. This process is denoted by $q_N(\cdot) \colon \mathbb{R}^{m \times n} \rightarrow \mathbb{R}_{N}^{m \times n}$, where $\mathbb{R}_{N}: \{\mathcal{T}[i] \in \mathbb{R} | 0 \leq i < 2^N \}$.

LoftQ (LoRA-Fine-Tuning-aware Quantization), introduced by \citet{li2023loftq}, is a method that addresses the performance degradation commonly observed when applying quantization alongside LoRA fine-tuning. LoftQ mitigates this issue by iteratively refining the quantized weights and their low-rank approximations, thus reducing the discrepancy between the quantized model and its full-precision counterpart. This process enhances the initialization for LoRA fine-tuning, resulting in improved performance on downstream tasks, particularly in low-bit quantization settings.

\begin{table*}[t]
    \centering
    \resizebox{0.86\textwidth}{!}{
    \begin{tabular}{@{}llcccccccc@{}}
        \toprule
        Pruning Rate &  & BoolQ & PIQA & HellaSwag & WinoGrande & ARC-e & ARC-c & OBQA & Memory (GB)  \\
        \midrule
        \multirow{3}{*}{Rate = 20\%} 
         & LoRA  & 63.30 & \textbf{76.82} & \textbf{68.68} & \textbf{63.38} & 63.76 & 37.11 & \textbf{40.60} & 35.06 \\
         & LoftQ & \textbf{67.77} & 76.55 & 68.03& 61.80 & 64.06 & 38.65 & 40.00 & 31.16 \\
         & LoftQ* & 66.42 & 76.66 & 67.96 & 60.93 & \textbf{65.31} & \textbf{38.40} & 40.00 & \textbf{24.52}  \\
         \bottomrule
        \end{tabular}}
\caption{Overall performance of seven benchmarks
for the different fine-tuning configurations. LoftQ denotes using uniform 8-bit quantization for all layers, whereas
LoftQ* indicates using mixed-precision quantization.}
\label{tab:Motivating Example}
\end{table*}

\subsection{The Motivating Example}\label{motivation}

Efficient fine-tuning of LLMs on resource-constrained devices requires effective model compression and fine-tuning techniques. We explored the combination of pruning, quantization, and LoRA fine-tuning to achieve this goal. Structural pruning reduces model size by removing less important parameters, but due to the varying importance of different layers \cite{zhang2023adalora}, it often results in uneven pruning across layers. This uneven pruning leads to a complex and unbalanced network structure, and standard quantization and fine-tuning processes, which typically apply a uniform configuration across all layers, may not be the optimal choice. 

To explore better configurations, we employed mixed-precision quantization, assigning different computational resources and complexities to different layers, with the aim of allowing more important layers to learn with finer granularity.

We conducted experiments using the llama-7b model with a pruning rate of 20\%. The pruning was performed using the optimal strategy determined by LLM-Pruner, leveraging its fine-tuning dataset and various hyperparameter configurations. The only differences among the settings are as follows: 1) using LoRA with a uniform 16-bit configuration across all layers; 2) initializing with LoftQ and applying a uniform 8-bit quantization across all layers; and 3) initializing with LoftQ and applying a mixed-precision setting, where each layer is randomly assigned a quantization bit-width of either 4 or 8 bits. 

As shown in Table \ref{tab:Motivating Example}, Configuration 3 significantly reduced overall computational complexity and decreased memory usage by approximately 30.1\% compared to Configuration 1, while still achieving better performance on certain tasks. Compared to Configuration 2, Configuration 3 reduced overall computational complexity too, with performance varying across tasks. This result suggests that for pruned models, a uniform fine-tuning configuration does not necessarily represent the performance ceiling, and by more finely allocating computational resources, the performance of compressed models can be further enhanced.

\section{Method}
\subsection{Problem Formalization}\label{3.1}
Inspired by the motivating example in Section \ref{motivation}, fine-grained mixed-precision quantization provides a promising approach to explore the upper bounds of performance for pruned models during efficient fine-tuning.

Consider a pruned LLM with $\mathit{L}$ layers. For each layer $i$ ($i \in \{1, 2, \dots, \mathit{L}\}$), we assign a mixed-precision quantization configuration $q_i$, which specifies the bit-width used to represent the weights and activations of the LoRA matrices during fine-tuning. The overall configuration across all layers is represented as a vector $\mathbf{q} = [q_1, q_2, \dots, q_L]$.

Let $\mathit{P}(\mathbf{q})$ denote the performance of the model on a downstream task under the quantization configuration $\mathbf{q}$, and let $\mathit{M}(\mathbf{q})$ denote the corresponding memory consumption. Our objective is to optimize the trade-off between these two conflicting goals: minimizing memory consumption while maximizing performance.

The search space $\mathbb{S}$ represents the set of all possible quantization configurations, structured as a tensor with dimensions $\mathit{L} \times \mathit{n}^\mathit{L}$, where $\mathit{n}$ is the number of quantization options. Each element in this tensor corresponds to a specific configuration that can be applied to the model layers.

It is important to note that the optimal quantization configuration $\mathbf{q}^*$ can vary significantly depending on the specific model architecture and the task it is being fine-tuned for. Different models have different sensitivities to quantization at various layers, and the task will require the model weights to change in different directions. As a result, there is no one-size-fits-all configuration, and the challenge lies in efficiently identifying the best configuration for each unique model-task combination.

This problem can be formulated as a multi-objective optimization problem, where the goal is to find the suitable configuration $\mathbf{q}^*$ that minimizes the following objective function:
\begin{equation}
\mathbf{q}^* = \arg \min_{\mathbf{q} \in \mathbb{S}} \left( \mathit{M}(\mathbf{q}) - \lambda \mathit{P}(\mathbf{q}) \right).
\label{eq:object_function}
\end{equation}
where $\lambda$ is a trade-off parameter that controls the relative importance of memory consumption versus task performance. By adjusting $\lambda$, we can prioritize memory efficiency or performance, ensuring that the selected configuration aligns with the desired optimization outcome.

The size of this search space $\mathbb{S}$ grows exponentially with $\mathit{L}$ and $\mathit{n}$, making exhaustive search computationally prohibitive. Evaluating the performance and memory consumption for each configuration in this vast space would require an immense amount of time and resources, particularly for large models with numerous layers. This presents a significant challenge in identifying the optimal quantization configuration.

Given the impracticality of exhaustive search, efficient optimization techniques are essential to navigate this search space and identify near-optimal configurations without evaluating every possible option. In the following sections, we address this challenge by introducing a performance model combined with Pareto optimality to guide the search process and efficiently explore the most promising regions of the configuration space.

\subsection{Performance Model and Pareto Optimality}\label{3.2}

\textbf{Performance Model.} To alleviate the computational burden of exhaustive search, we introduce the performance model. This model is designed to predict
the performance of unseen configurations based on a subset of mixed-precision quantization configurations that have undergone actual fine-tuning.

There are several potential architectures for this model. One approach is to use a data-driven method, such as a Multi-Layer Perceptron (MLP) or Transformer model. In this case, it is crucial to consider the encoding of input configurations, ensuring that the model effectively captures the relationships between different layers and quantization precisions. To enhance generalization across various models and tasks, meta-learning techniques can be employed, allowing the performance model to adapt quickly without requiring retraining from scratch for each new model or task. However, this approach may necessitate more manual intervention in the design of the encoding scheme.

Alternatively, a Bayesian approach using Gaussian Processes (GP) can be implemented. This method involves continuously updating the model with new data points and using the GP's uncertainty estimates to identify the next most promising configuration to evaluate. The GP model offers a balance between exploration and exploitation, guiding the search process by focusing on areas of the configuration space that are likely to yield the best trade-off between memory consumption and performance.

The performance model's objective is to minimize the difference between predicted and actual performance, which can be formalized as minimizing the loss function:
\begin{equation}
\mathcal{L}_{\text{perf}} = \sum_{j=1}^{N} \left| \mathcal{P}(\mathbf{q}_j) - \hat{\mathcal{P}}(\mathbf{q}_j) \right|,
\end{equation}
where $\mathcal{P}(\mathbf{q}_j)$ represents the actual performance for configuration $\mathbf{q}_j$, $\hat{\mathcal{P}}(\mathbf{q}_j)$ is the predicted performance, and $N$ is the number of training samples.

\textbf{Pareto Optimality.} In our optimization, we consider two conflicting objectives: minimizing memory consumption $\mathit{M}(\mathbf{q})$ and maximizing downstream task performance $\mathit{P}(\mathbf{q})$. A quantization configuration $\mathbf{q}_1$ is said to dominate another configuration $\mathbf{q}_2$ if:
\begin{equation}
\mathit{M}(\mathbf{q}_1) \leq \mathit{M}(\mathbf{q}_2) \quad \text{and} \quad \mathit{P}(\mathbf{q}_1) \geq \mathit{P}(\mathbf{q}_2),
\end{equation}
with at least one strict inequality. The set of all non-dominated configurations forms the Pareto frontier, represented by the Pareto optimal set $\mathbb{Q}^*$:
\begin{equation}
\mathbb{Q}^* = \left\{\mathbf{q} \in \mathbb{S} \mid \nexists \, \mathbf{q}' \in \mathbb{S} \, \phi(\mathbf{q}', \mathbf{q}) \right\},
\end{equation}
where $\phi(\mathbf{q}', \mathbf{q})$ represents the condition that $\mathit{M}(\mathbf{q}') \leq \mathit{M}(\mathbf{q})$ and $\mathit{P}(\mathbf{q}') \geq \mathit{P}(\mathbf{q})$.

By evaluating the configurations on this Pareto frontier, we identify those that achieve a balanced trade-off between memory consumption and performance. The final selection of the configuration is then made by applying the objective function previously defined in Equation \ref{eq:object_function}, adjusting the trade-off parameter $\lambda$ to prioritize either memory efficiency or performance, depending on the specific requirements of the application.

\begin{figure}[htbp]
  \centering
  \includegraphics[width=0.4\textwidth]{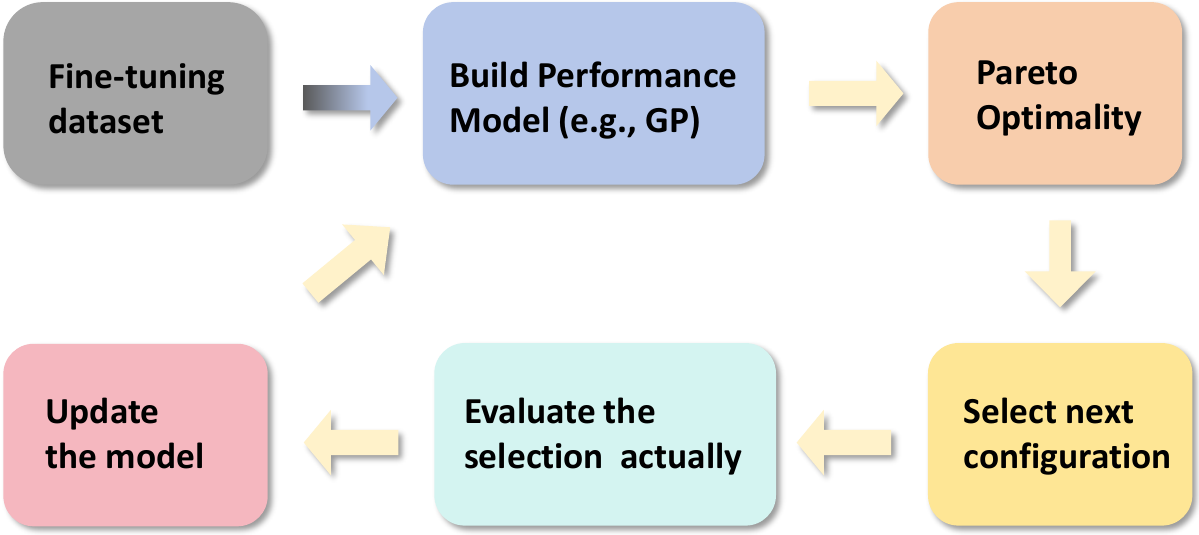}
    \caption{The workflow of AutoMixQ begins with a fine-tuning dataset, which is used to build a performance model. The Pareto frontier is established based on the model's predictions and known data, and a configuration that best fits the objective function is selected. This configuration is then fine-tuned on the LLM, and the fine-tuning results are used to update the performance model. This cycle of prediction, selection, evaluation, and updating continues until the Pareto frontier stabilizes or a predefined iteration limit is reached.}
    \label{fig:optimization_flow}
\end{figure}

\subsection{The Workflow of AutoMixQ}

Building upon the problem formalization and the introduction of our key components, we now present the workflow of AutoMixQ, which integrates these elements into a cohesive optimization pipeline.

The workflow is outlined in the figure \ref{fig:optimization_flow}. The process begins with a fine-tuning dataset for a given model and task, which includes initial performance and memory consumption data under various mixed-precision quantization configurations. These initial data points, while not strictly necessary for models like Gaussian Processes (GP), provide a solid starting point for the iterative process.

The performance model, designed to estimate the performance of unseen configurations, is then trained on the available data. In our implementation, we utilize GP due to its ability to iteratively predict the next most promising configuration to evaluate, thus significantly reducing the computational cost and resource usage compared to purely data-driven methods.

Next, the model generates predictions for various configurations, and these predictions are passed through a Pareto optimization process. This process identifies a set of configurations on the Pareto frontier, where no other configurations can improve one objective (memory consumption or performance) without degrading the other. After selecting a configuration from the Pareto frontier using the objective function defined in Equation \ref{eq:object_function}, the model undergoes actual fine-tuning to obtain real performance and memory data. 

These new data points are then used to update the performance model, enhancing its prediction accuracy. This cycle of prediction, optimization, evaluation, and updating continues until the Pareto frontier stabilizes or a predefined iteration limit is reached.

This iterative approach allows AutoMixQ to efficiently navigate the configuration space, identifying near-optimal quantization configurations that strike a balance between memory efficiency and performance, ultimately saving significant time and computational resources. The iterative process employing GP as the performance model is detailed in Section \ref{pareto_train}.


\section{Experiments}
\subsection{Experimental Setup}
\label{4.1}
\noindent{\textbf{LLMs and Benchmarks.}}
To demonstrate how AutoMixQ performes on different model, we test it on three open source large language models: LLaMA-7B \cite{touvron2023llama}, LLaMA-13B \cite{touvron2023llama} and Vicuna-7B \cite{zheng2024judging}, and specific version is stated in the Appendix \ref{LLM}. We conduct these LLMs on zero-shot classification tests for commonsense reasoning datasets, including BoolQ \cite{clark2019boolq}, PIQA \cite{bisk2020piqa}, HellaSwag \cite{zellers2019hellaswag}, WinoGrande \cite{sakaguchi2021winogrande}, ARC-easy \cite{clark2018think}, ARC-challenge \cite{clark2018think}, and OpenbookQA \cite{mihaylov2018can}.

\noindent{\textbf{Software and hardware configuration.}}
We utilize the following configurations: \textit{PyTorch} version 2.1.2, \textit{BitsandBytes} library version 0.43.1, \textit{Transformers} library version 4.41.0, \textit{PEFT (Parameter-Efficient Fine-Tuning)} library version 0.11.1, \textit{Optuna} library version 3.6.1, \textit{CUDA} version 12.4, \textit{GPU:} NVIDIA L20 GPU with 48GB of memory. \textit{Operating System:} Ubuntu.

\noindent \textbf{Baselines.} To demonstrate the effectiveness of our method, we compare it against the following three baselines: 
\begin{itemize}
    \item \textit{Original Model:} The original model without pruning. 
    \item \textit{16-bit LoRA:} The pruned model with 16-bit precision, fine-tuned using the LoRA method.
    \item \textit{8-bit LoftQ:} The pruned model quantized to 8-bit precision, fine-tuned using the LoftQ method.
\end{itemize}

\noindent \textbf{Implementation Details.}
The pruning method follows LLM-Pruner \citep{ma2023llm}, and the dataset uses 50k publicly available samples from the Alpaca \cite{alpaca}. All experiments were conducted with a LoRA matrix rank of 8, and LoftQ initialization with one iteration. We utilized BitsandBytes for quantization configuration. For 4-bit quantization, we employed NF4 \cite{dettmers2024qlora}, and since 2-bit quantization does not reduce memory usage \cite{li2023loftq}, each layer's quantization configuration only considered 4-bit and 8-bit options. More detailed hyperparameter settings can be found in Appendix \ref{appendix:hyperparams}.

\subsection{\textbf{Details of The Optimization Workflow.}}
\label{pareto_train}

\begin{figure}[h]
    \centering
    \begin{subfigure}[b]{0.45\textwidth}
        \centering
        \includegraphics[width=\textwidth]{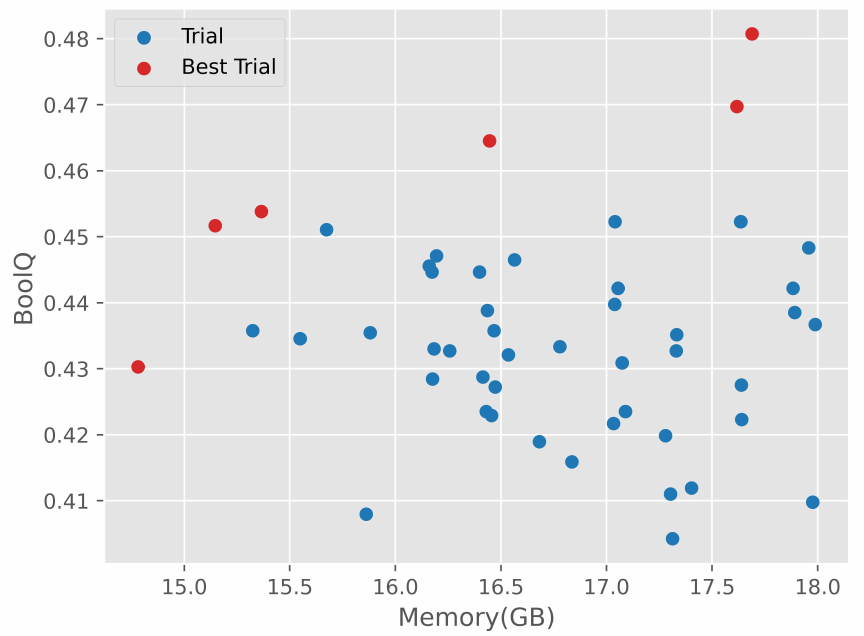}
        \caption{Pareto-front scatter plot for BoolQ}
        \label{fig:sub1}
    \end{subfigure}
    \hfill 
    \begin{subfigure}[b]{0.45\textwidth}
        \centering
        \includegraphics[width=\textwidth]{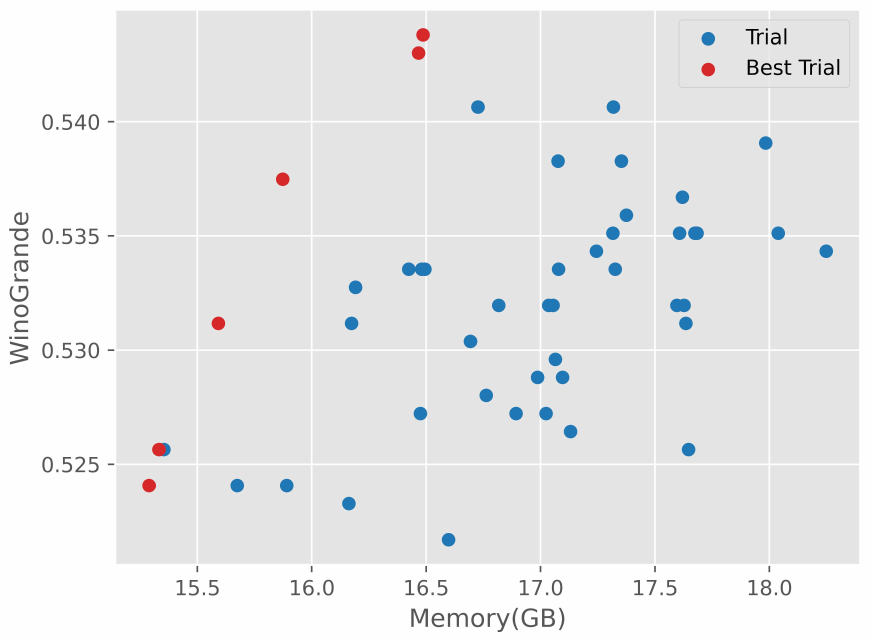}
        \caption{Pareto-front scatter plot for WinoGrande}
        \label{fig:sub2}
    \end{subfigure}
    \caption{Pareto-front scatter plots for BoolQ and WinoGrande with 50 data points. The red points indicate the non-dominated configurations within the Pareto frontier.}
    \label{fig:test}
\end{figure}

We illustrated the optimization process, memory, and time footprint of AutoMixQ using a 50\% parameter pruning rate on llama-7b as an example. We fine-tuned 10 sets of configurations as the initialization for the Gaussian Process (GP) (this is not mandatory; in other experiments, we found that starting from scratch, a good configuration could be found in about 10 iterations). In each configuration, the quantization precision for all model layers was randomly selected between 4-bit and 8-bit. On average, obtaining data for each initialization took approximately 25 minutes. We set the total number of iterations for AutoMixQ to 40 (resulting in 50 data points for constructing the Pareto front) to ensure the best configuration was found. The entire process took approximately 16.5 hours. During AutoMixQ iterations, GP required around 7s to suggest the next configuration, while the prediction process and Pareto frontier construction consumed approximately 187MB memory. In Figure \ref{fig:test}, we present the optimization results for BoolQ and Winograd. More detailed processes and results for other benchmarks are provided in Appendix \ref{appendix:pareto}.


\subsection{Main Results}
\label{MainResults}
We tested models with different pruning rates across the benchmarks mentioned earlier to compare the performance of our method against the baselines. We report the performance and peak memory usage of the models under different fine-tuning methods. Since the LLaMA paper did not provide specific test prompts, we used the open prompts created by \citet{eval-harness} for the benchmarks. The results for LLaMA-7B and Vicuna-7B are shown in Table \ref{table:result-comp}, for LLaMA-13B in Appendix \ref{res-llama13b}. It would be interesting to test our method on larger models, such as 70B or higher, but due to hardware limitations, we have not yet conducted these experiments. Additionally, we present the generation quality of the fine-tuned models obtained using the three methods in Appendix~\ref{llama7b_generation}. These results demonstrate the superiority of AutoMixQ.

\begin{table*}[t]
    \centering
    \resizebox{0.86\textwidth}{!}{\begin{tabular}{@{}llcccccccc@{}}
        \toprule
        & &Method & BoolQ & PIQA & HellaSwag & WinoGrande & ARC-e & ARC-c & OBQA \\
        \midrule
        \multirow{11}{*}{\parbox{2cm}{LLaMA-7B}} & {\parbox{1.8cm}{Rate = 0\%}} & w/o tuning& 73.09 & 78.35 & 72.98 & 67.09 & 67.42 & 41.38 & 42.40 \\
        \cmidrule{2-10}
         & \multirow{3}{*}{Rate = 20\%} 
         & LoRA  & 63.30(35.06) & 76.82(35.06) & \textbf{68.68}(35.06) & \textbf{63.38}(35.06) & 63.76(35.06) & 37.11(35.06) & 40.60(35.06) \\
         & & LoftQ & 67.77(31.16) & 76.55(31.16) & 68.03(31.16) & 61.80(31.16) & 64.06(31.16) & 38.65(31.16) & 40.00(31.16) \\
         & & AutoMixQ & \textbf{67.98}(\textbf{24.17}) & \textbf{77.01}(\textbf{22.83}) & 68.18(\textbf{23.32}) & 62.73(\textbf{24.42}) & \textbf{66.16}(\textbf{23.48}) & \textbf{38.91}(\textbf{24.42}) & \textbf{40.80}(\textbf{24.14}) \\
        \cmidrule{2-10}
         & \multirow{3}{*}{Rate = 30\%} 
         & LoRA  & 62.45(31.38) & 74.37(31.38) & \textbf{63.14}(31.38) & \textbf{61.96}(31.38) & \textbf{59.22}(31.38) & 33.70(31.38) & \textbf{39.60}(31.38) \\
         & & LoftQ & 58.96(27.94) & 71.22(27.94) & 58.10(27.94) & 58.88(27.94) & 52.19(27.94) & 32.34(27.94) & 38.40(27.94) \\
         & & AutoMixQ & \textbf{66.21}(\textbf{20.26}) & \textbf{74.43}(\textbf{22.14}) & 61.14(\textbf{22.15}) & 61.40(\textbf{20.63}) & 58.12(\textbf{21.67}) & \textbf{34.47}(\textbf{22.15}) & 39.00(\textbf{22.15}) \\
        \cmidrule{2-10}
         & \multirow{3}{*}{Rate = 50\%} 
         & LoRA  & 43.76(23.89) & 68.88(23.89) & 44.85(23.89) & 50.99(23.89) & 45.20(23.89) & 28.75(23.89) & 34.60(23.89) \\
         & & LoftQ & 45.14(20.40) & 68.34(20.40) & 44.39(20.40) & 52.96(20.40) & 43.86(20.40) & 29.01(20.40) & 35.80(20.40) \\
         & & AutoMixQ & \textbf{48.07}(\textbf{17.69}) & \textbf{68.98}(\textbf{16.23}) & \textbf{44.89}(\textbf{17.35}) & \textbf{54.45}(\textbf{16.41}) & \textbf{45.28}(\textbf{16.46}) & \textbf{29.35}(\textbf{16.41}) & \textbf{36.40}(\textbf{16.78}) \\
        \midrule
        \multirow{12}{*}{\parbox{2cm}{Vicuna-7B}} & {\parbox{1.8cm}{Rate = 0\%}} & w/o tuning &75.69 & 77.75 & 71.06 & 67.80 & 69.07 & 40.78 & 42.20 \\
        \cmidrule{2-10}
         & \multirow{3}{*}{Rate = 20\%} 
         & LoRA  & 57.77(35.25) & 77.56(35.25) & 67.16(35.25) & \textbf{63.14}(35.25) & 67.30(35.25) & 37.71(35.25) & 40.40(35.25) \\
         & & LoftQ & 57.95(31.11) & 76.82(31.11) & 66.42(31.11) & 62.51(31.11) & 66.62(31.11) & 37.37(31.11) & 40.60(31.11) \\
         & & AutoMixQ & \textbf{59.75}(\textbf{22.56}) & \textbf{77.59}(\textbf{20.88}) & \textbf{67.21}(\textbf{23.95}) & 62.98(\textbf{23.47}) & \textbf{67.45}(\textbf{23.47}) & \textbf{37.85}(\textbf{23.99}) & \textbf{41.20}(\textbf{25.22}) \\
        \cmidrule{2-10}
         & \multirow{3}{*}{Rate = 30\%} 
         & LoRA  & \textbf{58.81}(31.83) & 74.37(31.83) & 60.70(31.83) & \textbf{60.62}(31.83) & 59.01(31.83) & 33.79(31.83) & 38.80(31.83) \\
         & & LoftQ & 53.85(27.82) & 74.76(27.82) & 60.65(27.82) & 60.06(27.82) & 59.72(27.82) & 34.30(27.82) & 38.20(27.82) \\
         & & AutoMixQ & 56.76(\textbf{23.06}) & \textbf{75.90}(\textbf{24.02}) & \textbf{60.99}(\textbf{21.68}) & 60.37(\textbf{20.69}) & \textbf{60.81}(\textbf{21.68}) & \textbf{34.70}(\textbf{20.65}) & \textbf{39.80}(\textbf{20.65}) \\
        \cmidrule{2-10}
         & \multirow{3}{*}{Rate = 50\%} 
         & LoRA  & 59.51(24.55) & 66.87(24.55) & 43.18(24.55) & 52.01(24.55) & 48.40(24.55) & 26.45(24.55) & 34.00(24.55) \\
         & & LoftQ & 59.51(20.43) & 67.90(20.43) & 43.30(20.43) & 50.83(20.43) & 48.82(20.43) & 27.73(20.43) & 34.60(20.43) \\
         & & AutoMixQ & \textbf{60.76}(\textbf{17.24}) & \textbf{68.28}(\textbf{15.76}) & \textbf{43.72}(\textbf{15.76}) & \textbf{52.88}(\textbf{15.50}) & \textbf{49.66}(\textbf{15.49}) & \textbf{27.98}(\textbf{16.90}) & \textbf{35.80}(\textbf{15.49}) \\
        \bottomrule
    \end{tabular}}
    \caption{Zero-shot performance and peak memory usage on LLaMA-7B (top) and Vicuna-7B (bottom) with varying pruning rates. LoRA uses 16-bit quantization, LoftQ uses 8-bit quantization, and AutoMixQ employs mixed-precision quantization. ‘Bold’ indicates the best performance and memory usage at each pruning rate. The performance is reported in percentage (\%), and the values in parentheses represent peak memory usage(in GB).}
    \label{table:result-comp}
\end{table*}

\noindent{\textbf{Different Tasks Analysis.}}
Across various tasks, AutoMixQ consistently demonstrated significantly lower memory usage compared to LoRA and LoftQ, with an average reduction of 33\% relative to LoRA and 25\% relative to LoftQ. Despite this, AutoMixQ consistently outperforms LoftQ and, in most cases, surpasses LoRA as well. For example, on the ARC-c dataset, known for its challenging multiple-choice questions that require deep reasoning, AutoMixQ achieves the best performance when fine-tuning LLaMA-7B, outperforming both LoftQ and LoRA. At a 20\% pruning rate, AutoMixQ improves accuracy by 1.8\% over LoRA while reducing memory usage by 30.2\%. The results on BoolQ and WinoGrande further demonstrate AutoMixQ's advantages. In these tasks, which involve binary questions and sentence completion respectively, AutoMixQ not only achieves the highest accuracy compared to LoftQ and LoRA but also significantly reduces memory usage. On BoolQ, with a 20\% pruning rate, AutoMixQ outperforms LoRA by 4.68\% while reducing memory usage by 30.9\%. At a 50\% pruning rate on WinoGrande, AutoMixQ increases accuracy by 3.46\% over LoRA while reducing memory usage by 38.9\%. AutoMixQ also demonstrates clear advantages on Vicuna-7B, indicating its ability to effectively balance the trade-off between maintaining accuracy and minimizing memory usage—a critical factor for deploying models in resource-constrained environments. This makes AutoMixQ an ideal choice for scenarios where memory resources are limited but high accuracy is still required.

\noindent{\textbf{Different Pruning Rates Analysis.}}
When analyzing the impact of different pruning rates, AutoMixQ continued to achieve the best balance between accuracy and memory efficiency. Particularly at a 50\% pruning rate, AutoMixQ maintains excellent performance while significantly reducing memory usage, demonstrating its robustness. For instance, in the ARC-e task with LLaMA-7B at a 50\% pruning rate, AutoMixQ matches LoRA in performance but reduces memory usage by 31.1\%. Compared to LoftQ, AutoMixQ improves performance by 1.42\% while reducing memory usage by 19.3\%. Additionally, in the OBQA task, AutoMixQ outperforms LoRA by 1.8\% while reducing memory usage by 30\%. These results highlight AutoMixQ's ability to maintain high accuracy even under aggressive compression conditions. Furthermore, in the ARC-e task with Vicuna-7B at a 50\% pruning rate, AutoMixQ improves accuracy by 1.26\% over LoRA while reducing memory usage by 36.9\%. Although the performance improvement over LoftQ is less than 1\%, the memory usage reduction remains significant. This demonstrates the versatility of the AutoMixQ approach across different models, further proving AutoMixQ's capability to approach the upper bounds of model performance.

\noindent{\textbf{Summary.}}
The results show that, under the same fine-tuning data and settings, AutoMixQ outperforms both LoftQ and LoRA in terms of performance and memory efficiency, especially in tasks requiring complex reasoning and at higher pruning rates. It is important to note that due to its self-adjusting nature, AutoMixQ is versatile and can be applied across different pruning methods, models, and tasks. This makes it an effective solution for deploying large models in environments with limited hardware resources, as generalization, accuracy, and resource usage are critical issues in real-world applications.

\subsection{Ablation Study}
In this part, we use LLaMA-7B with 20\% pruning rate and a configuration obtained by the AutoMixQ method to conduct ablation experiments and all results are shown in Table \ref{tab:Ablation}. 

\noindent{\textbf{Dtype of 4-bit.}} Performance was compared using different 4-bit data types. The performance of NF4 and FP4 is different, which shows that the AutoMixQ method has no dependence on the quantization type.

\noindent{\textbf{Adapter initialization Method.}} Different initial weights for the adapter layers are tested LoftQ, Gaussian, PiSSA \cite{meng2024pissa}. There is no obviously dominant initialization method, but it shows that the AutoMixQ method is effective in different initialization methods.

\noindent{\textbf{Adapter Iteration Count.}} In the performance comparison of the LoftQ method matrix initialization iterations 1, 2, and 4, the performance decreases with more iterations, which shows that using low-rank matrix fitting residuals to reduce quantization errors does not necessarily improve model performance. LoftQ recommends a setting of 1 too.

\begin{table*}[h]
\centering
\resizebox{0.87\linewidth}{!}{
\begin{tabular}{l||cc||ccc||ccc||cc}
\toprule
\multirow{2}{*}{\textbf{Benchmark}} & \multicolumn{2}{c||}{\textbf{Dtype of 4-bit}} & \multicolumn{3}{c||}{\textbf{Adapter Initialization Method}} & \multicolumn{3}{c||}{\textbf{Adapter Iteration Count}} & \multicolumn{2}{c}{\textbf{Importance Estimation}} \\
\cmidrule(lr){2-3} \cmidrule(lr){4-6} \cmidrule(lr){7-9} \cmidrule(lr){10-11}
 & \textbf{NF4} & \textbf{FP4} & \textbf{LoftQ} & \textbf{Gaussian} & \textbf{PiSSA} & \textbf{iter=1} & \textbf{iter=2} & \textbf{iter=4} & \textbf{Element\textsuperscript{1}} & \textbf{Element\textsuperscript{2}} \\
\midrule
ARC-e       & \textbf{65.49} & 62.84 & \textbf{65.49} & 64.77 & 64.44 & \textbf{65.49} & 64.31 & 64.18 & \textbf{65.49} & 62.50 \\
ARC-c       & \textbf{38.99} & 36.77 & \textbf{38.99} & \textbf{38.99} & 38.40 & \textbf{38.99} & 38.05 & 38.14 & \textbf{38.99} & 37.80 \\
WinoGrande  & 61.40 & \textbf{63.22} & 61.40 & \textbf{61.96} & 61.48 & \textbf{61.40} & 60.46 & 60.69 & \textbf{61.40} & 59.43 \\
OBQA        & \textbf{40.20} & 39.80 & 40.20 & 39.00 & \textbf{40.40} & \textbf{40.20} & 39.40 & 39.60 & \textbf{40.20} & 38.60  \\
BoolQ       & \textbf{67.22} & 66.48 & 67.22 & 64.43 & \textbf{68.20} & 67.22 & \textbf{67.55} & 66.85 & \textbf{67.22} & 65.44 \\
PIQA        & \textbf{76.82} & \textbf{76.82} & \textbf{76.82} & 76.44 & 76.39 & \textbf{76.82} & 76.44 & 76.55 & \textbf{76.82} & 76.39 \\
HellaSwag   & \textbf{67.97} & 67.88 & 67.97 & 67.80 & \textbf{68.01} & \textbf{67.97} & \textbf{67.97} & 67.93 & \textbf{67.97} & 66.93 \\
\bottomrule
\end{tabular}
}
\caption{Performance comparison of ablation studies on seven tasks at 20\% pruning rate on LLaMA-7B. It appears that AutoMixQ captures potential resource allocations without relying on other settings. Results are reported in percentage (\%).}
\label{tab:Ablation}
\end{table*}
\noindent{\textbf{Importance Estimation Method}} Tested the pruning process parameter importance estimation method. The results compare the first-order (Element\textsuperscript{1}) and second-order (Element\textsuperscript{2}) Taylor approximations for estimating the importance of each parameter. The results show that Element\textsuperscript{1} provides better performance than Element\textsuperscript{2} for all benchmarks. Although higher-order derivatives can theoretically provide more precise tuning, no performance improvement was observed, further illustrating the complexity of the layer relationships in LLMs.

\section{Related Work}
\subsection{Efficient Compression of LLMs}
LLM-Pruner \cite{ma2023llm} uses structured pruning to eliminate non-essential interconnected structures by leveraging gradient information. This technique enables compressed models to maintain good performance across multiple tasks with basic fine-tuning. \citet{santacroce2023matters} proposes Globally Unique Movement (GUM), a novel pruning technique focusing on the sensitivity and uniqueness of LLMs' network components. GUM prunes neurons that uniquely contribute to the model output and are sensitive to loss changes, thus preserving high accuracy. This method optimizes the trade-off between information retention and computational efficiency. SparseGPT \cite{frantar2023sparsegpt} is a pruning method that transforms the process into a series of large-scale sparse regression problems, solvable through Hessian matrix inversion without retraining. It efficiently prunes large models to high sparsity in a single step while maintaining high accuracy. Wanda \cite{sun2023simple} prunes LLMs by selectively removing weights based on their sizes and input activations, adaptively adjusting sparsity levels to reduce more than half without sacrificing accuracy. Quantization-Aware Training (QAT) combines quantization with full model fine-tuning to adapt models for downstream tasks \citep{peri2020deploying, liu2023llm}. Although QAT is effective, it requires substantial computational resources, such as gradient calculations and optimization states, and it complicates the gradient computation for quantized weights. However, by leveraging LoRA, these challenges can be bypassed during task adaptation. Post-Training Quantization (PTQ) frameworks, such as GPTQ and SmoothQuant \citep{frantar2022gptq, xiao2023smoothquant}, use a small subset of training data to calibrate high-precision models, enabling the generation of task-specific quantized models without the need for gradient backpropagation. This makes PTQ more cost-efficient than QAT, although it generally results in lower accuracy.


\subsection{Parameter Efficient Fine-Tuning}
\citet{houlsby2019parameter} introduce a transfer learning method incorporating adapter modules into pre-trained Transformer models, efficiently handling various NLP tasks with few additional parameters while achieving performance comparable to full fine-tuning. LLM-Adapters \cite{hu2023llm} integrate small adapters with few extra parameters into LLMs for efficient fine-tuning, allowing smaller models to perform as well as larger ones on specific tasks. Unlike the serial approach of adapters, low-rank adaptation (LoRA) \cite{hu2021lora} uses a parallel method to insert trainable rank decomposition matrices into each layer of the model's architecture. LoRA adds trainable matrices to each layer while keeping the pre-trained weights unchanged, reducing the number of trainable parameters and making model adaptation faster and less resource-intensive. LoRA-FA \cite{zhang2023lora} freezes the projection-down weight of the LoRA layers and only updates the projection-up weight, reducing the memory requirements for fine-tuning. QLora \cite{dettmers2024qlora} combines low-rank adapters and quantized 4-bit weights for efficient LLM fine-tuning, significantly reducing GPU memory requirements while achieving performance comparable to full 16-bit fine-tuning. LoftQ \cite{li2023loftq} applies quantization and low-rank approximation alternately to achieve a good initialization for LoRA fine-tuning, mitigating the discrepancy between quantized and pre-trained weights, and enabling efficient fine-tuning of quantized models, particularly in challenging low-bit regimes.

\section{Conclusion}
We propose a novel fine-tuning framework, AutoMixQ, which organically combines pruning, quantization, and LoRA to achieve high-performance fine-tuned models under low-resource conditions. After formalizing the problem as an optimization problem, AutoMixQ employs an end-to-end automatic optimization flow, integrating lightweight performance models with Pareto optimality to rapidly self-adjust the quantization precision of model layers, thereby achieving the desired objectives. Evaluations conducted on popular benchmarks confirm that AutoMixQ delivers excellent memory efficiency and performance. Through an automatic optimization process, from the perspective of pruning, AutoMixQ achieves accuracy that more closely aligns with the original model; from the perspective of quantization, it ensures optimal resource allocation; and from the perspective of fine-tuning, it strikes a superior balance between performance and memory usage. Moreover, its self-adjusting nature ensures its broad applicability.

\bibliography{ref}

\appendix
\section{Version of LLMs}\label{LLM}
We provide the Hugging Face link of LLMs used in the experiment:
LLaMA-7B: \url{https://huggingface.co/baffo32/decapoda-research-llama-7B-hf}; Vicuna-7B: \url{https://huggingface.co/lmsys/vicuna-7b-v1.5}; LLaMA-13B: \url{https://huggingface.co/yahma/llama-13b-hf}

\section{Hyperparameters}
\label{appendix:hyperparams}
In the optimization of the pruned LLaMA-7B model, a comprehensive hyperparameter configuration was employed to ensure an optimal balance between model performance and computational efficiency. The model was fine-tuned with a learning rate of $3 \times 10^{-4}$, utilizing a batch size of 128, further divided into micro batches of 4 to manage memory constraints effectively. Sequences were standardized to a maximum length of 256 tokens, and a dropout of 0.05 was applied specifically to the LoRA layers targeting projections such as query, key, value, and output, alongside gate, down, and up projections. Quantization was dynamically applied at 4-bit and 8-bit levels according to layer requirements to optimize memory use without compromising computational accuracy. The training employed the paged AdamW optimizer with 32-bit precision, enhancing stability and efficiency. These settings were methodically tested and optimized through the Optuna framework to ensure robust model performance and resource utilization.

\section{Results of Optimization Workflow}
\label{appendix:pareto}
\begin{figure*}[h]
    \centering
    \begin{subfigure}{0.45\linewidth}
        \centering
        \includegraphics[width=\linewidth]{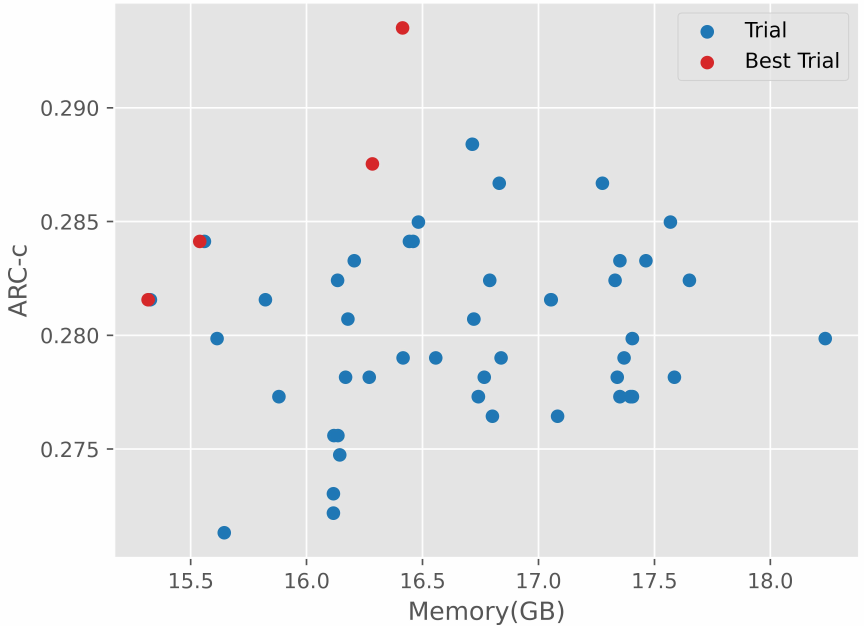}
        \caption{ARC-c}
        \label{fig:sub1}
    \end{subfigure}
    \begin{subfigure}{0.45\linewidth}
        \centering
        \includegraphics[width=\linewidth]{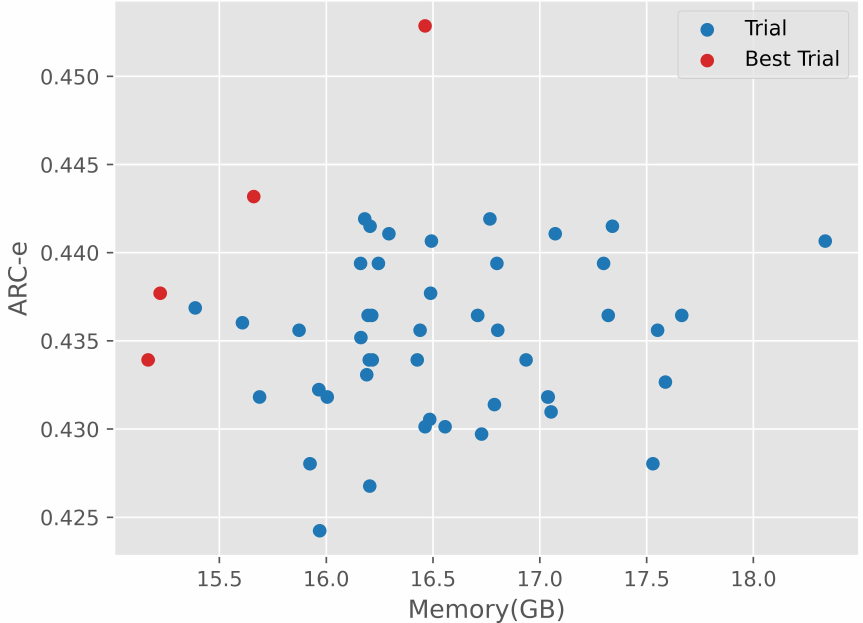}
        \caption{ARC-e}
        \label{fig:sub2}
    \end{subfigure}
    \begin{subfigure}[b]{0.45\linewidth}
        \centering
        \includegraphics[width=\linewidth]{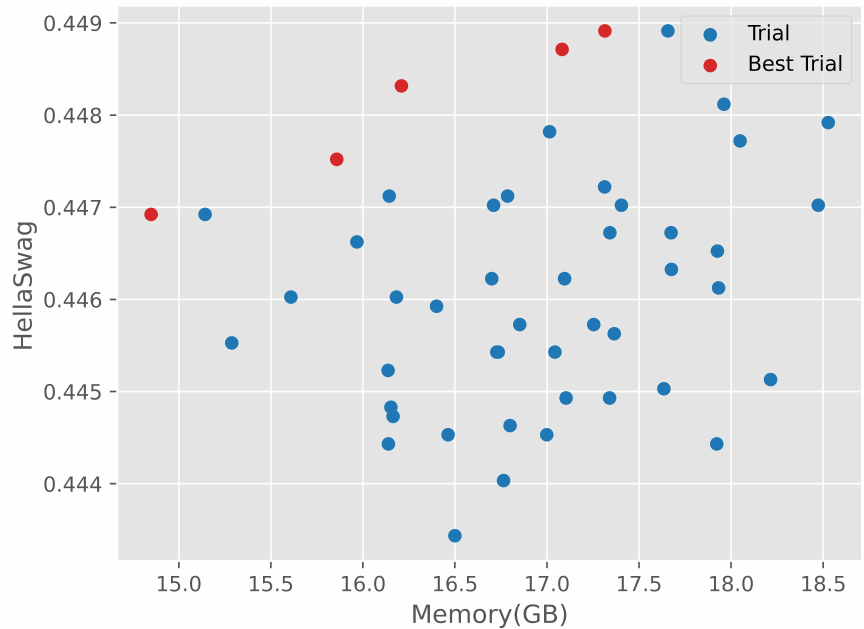}
        \caption{HellaSwag}
        \label{fig:sub3}
    \end{subfigure}
    \begin{subfigure}[b]{0.45\textwidth}
        \centering
        \includegraphics[width=\linewidth]{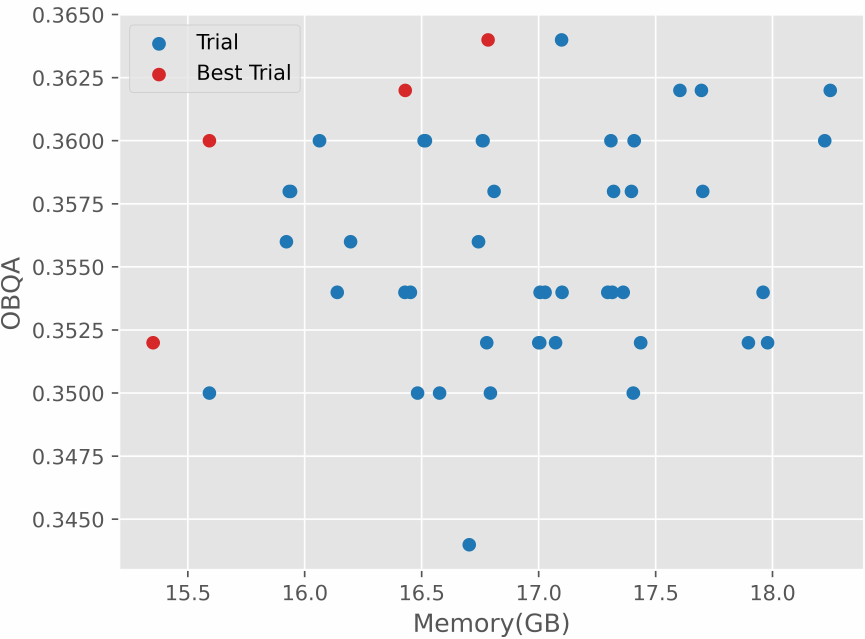}
        \caption{OBQA}
        \label{fig:sub4}
    \end{subfigure}
    \begin{subfigure}[b]{0.45\textwidth}
        \centering
        \includegraphics[width=\linewidth]{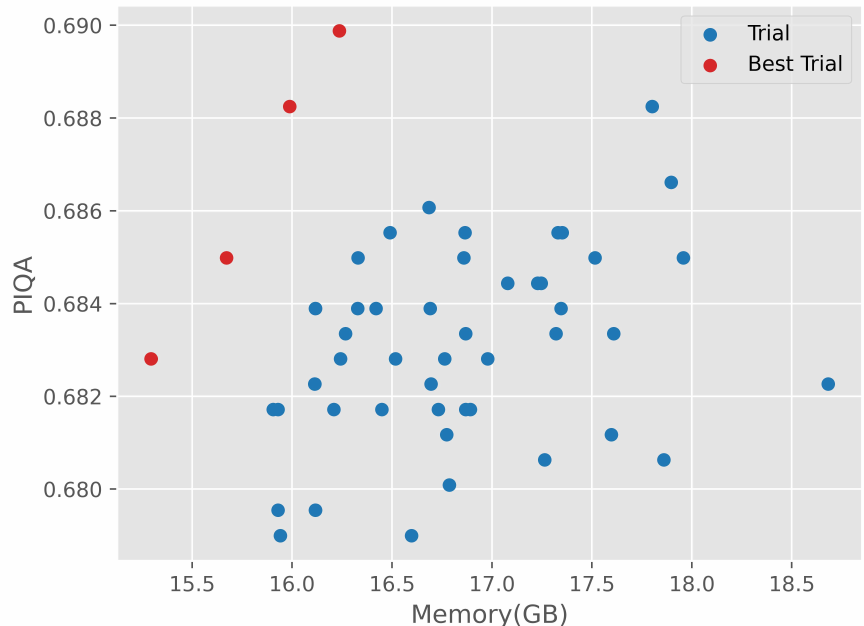}
        \caption{PIQA}
        \label{fig:sub5}
    \end{subfigure}
    
    \caption{Pareto-front scatter plots for different Downstream Tasks}
    \label{fig:combined_pareto_fronts}
\end{figure*}

In this section, we will use the LLaMA-7B model with 50\% pruning as our example to illustrate the Pareto optimization workflow, as shown in Figure \ref{fig:combined_pareto_fronts}

\section{Performance in LLaMA-13B}\label{res-llama13b}
We list the performance of the configuration described in Section \ref{4.1} for LLaMA-13B in Table \ref{llama13bperform}.

\section{Sample Generation result} \label{llama7b_generation}
\begin{figure*}[h]
    \centering
    \begin{subfigure}[h]{\textwidth}
        \centering
        \includegraphics[width=\textwidth]{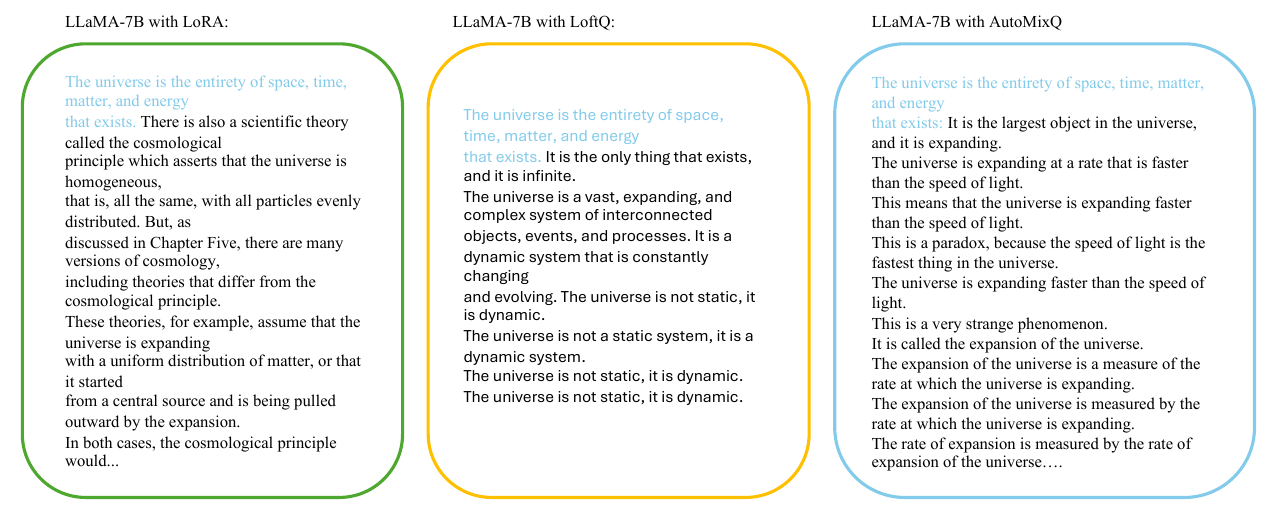}
        \caption{Sample generation for the LLaMA-7B model with a 20\% pruning rate.}
        \label{llama7bgen}
    \end{subfigure}
    \vspace{-10pt} 
    \begin{subfigure}[h]{\textwidth}
        \centering
        \includegraphics[width=\textwidth, height=0.3\textwidth]{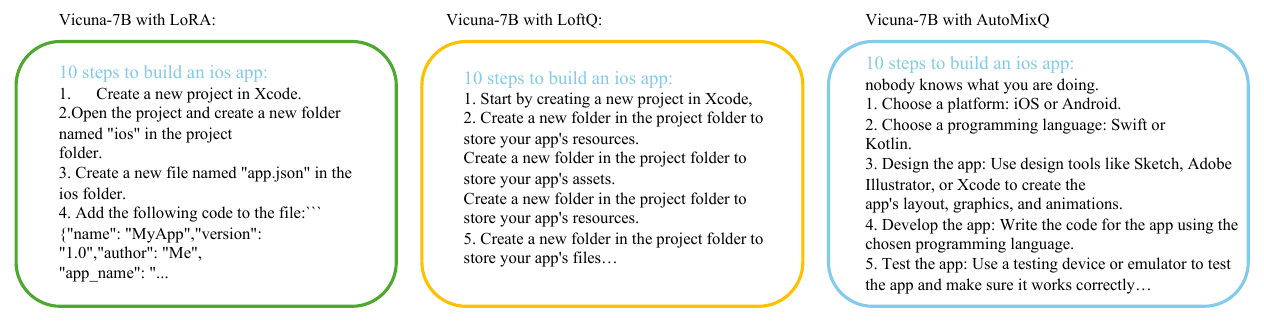}
        \caption{Sample generation for the Vicuna-7B model with a 20\% pruning rate and a maximum of 128 tokens.}
        \label{generation}
    \end{subfigure}
    \caption{Sample generation for LLaMA-7B and Vicuna-7B models with a 20\% pruning rate.}
    \label{fig:combined_generation_vertical}
\end{figure*}

In this section, we present the results on the LLaMA-7B model using the input prompt "The universe is the entirety of space, time, matter, and energy that exists." We compared the outcomes generated by LoRA, LoftQ, and our model. The results demonstrate that our model significantly outperforms LoftQ and closely approximates LoRA, indicating the high accuracy of our model. Figure \ref{llama7bgen} We also provide the generation result for Vicuna-7B in Figure \ref{generation}. For the Vicuna-7B model, our prompt was "10 steps to build an iOS app." We observed that the results generated by AutoMixQ were significantly better than those from LoftQ and, compared to LoRA, were more reasonable and accurate.

\begin{table*}[h]
    \centering
    \resizebox{\textwidth}{!}{
        \begin{tabular}{llccccccc}
            \toprule
            Pruning Rate & Recover & BoolQ & PIQA & HellaSwag & WinoGrande & ARC-e & ARC-c & OBQA \\
            \midrule
            {\parbox{1.8cm}{Rate = 0\%}} & w/o tuning & 68.50 & 79.11 & 76.21 & 70.09 & 74.58 & 44.54 & 42.20  \\
            \cmidrule{1-9}
            \multirow{3}{*}{\parbox{1.8cm}{Rate = 50\%}} 
            & LORA     & 61.93(41.32) & 71.38(41.32) & 53.36(41.32) & 53.59(41.32) & 29.95(41.32) & 53.11(41.32) & 38.00(41.32)  \\
            \cmidrule{2-9}
            & LoftQ    & 61.71(36.68) & 72.63(36.68) & 56.10(36.68) & 55.17(36.68) & 31.57(36.68) & 55.47(36.68) & 38.60(36.68)  \\
            \cmidrule{2-9}
            & AutoMixQ     & 61.80(30.53) & \textbf{73.23}(30.53) & \textbf{56.37}(30.53) & 55.09(31.45) & 31.48(30.53) & \textbf{55.80}(31.45) & \textbf{39.00}(30.58)  \\
            \bottomrule
        \end{tabular}}  
    \caption{Zero-shot performance and memory of LLaMA-13B in LoRA, LoftQ, and AutoMixQ. ‘Bold’ indicates the best performance at each pruning rate.  Reported in percentage (\%).}
    \label{llama13bperform}
\end{table*}

\end{document}